\documentclass{article}

     \usepackage[preprint]{neurips_2023}

\usepackage[utf8]{inputenc} 
\usepackage[T1]{fontenc}    
\usepackage{hyperref}       
\usepackage{url}            
\usepackage{booktabs}       
\usepackage{amsfonts}       
\usepackage{nicefrac}       
\usepackage{microtype}      
\usepackage{xcolor}         
\usepackage{graphicx}
\usepackage{subcaption}
\usepackage{enumitem}
\usepackage{inconsolata}

\usepackage{soul}

\title{Evaluating the Utility of Model Explanations \\ for Model Development}

\author{%
  Shawn Im \\
  UW-Madison\\
  \texttt{shawnim@cs.wisc.edu} \\
   \And
   Jacob Andreas \\
   MIT CSAIL \\
   \texttt{jda@mit.edu} \\
   \And
   Yilun Zhou \\
   MIT CSAIL \\
   \texttt{yilun@csail.mit.edu} \\
}

\begin{document}

\maketitle

\begin{abstract}
  One of the motivations for explainable AI is to allow humans to make better and more informed decisions regarding the use and deployment of AI models. But careful evaluations are needed to assess whether this expectation has been fulfilled. Current evaluations mainly focus on algorithmic properties of explanations, and those that involve human subjects often employ subjective questions to test human's perception of explanation usefulness, without being grounded in objective metrics and measurements. In this work, we evaluate whether explanations can improve human decision-making in practical scenarios of machine learning model development. We conduct a mixed-methods user study involving image data to evaluate saliency maps generated by SmoothGrad, GradCAM, and an oracle explanation on two tasks: model selection and counterfactual simulation. To our surprise, we did not find evidence of significant improvement on these tasks when users were provided with any of the saliency maps, even the synthetic oracle explanation designed to be simple to understand and highly indicative of the answer. Nonetheless, explanations did help users more accurately describe the models. These findings suggest caution regarding the usefulness and potential for misunderstanding in saliency-based explanations. 
\end{abstract}

\section{Introduction}

The factors that influence prediction in neural network models are often opaque. As a result, many methods have been developed to attempt to explain their predictions, including feature attribution explanations \citep{smilkov2017smoothgrad, selvaraju2017grad, lundberg2017unified, ribeiro2016should}, counterfactual explanations \citep{wachter2017counterfactual, dandl2020multi, ustun2019actionable} and concept-based explanations \citep{ghorbani2019towards, kim2018interpretability, koh2020concept, chen2020concept}. These explanations have been used for guiding human understanding of models but also have uses for pruning and data augmentation \citep{lee2018snip, kim2021co}. While they are primarily designed to provide information about the model's reasoning process in a human-interpretable way (e.g.\ such by highlighting important features), when and how they improve human decision-making about models on downstream tasks remains an ongoing topic of study. Previous work has evaluated their usefulness on tasks such as model debugging \citep{chen2022use}, verifying model reasoning \citep{lage2019evaluation}, and other understanding-related tasks \citep{pruthi2022evaluating, bansal2021does}; these studies have obtained positive results with certain explanations. 

In this paper, we focus on two tasks commonly encountered in the ML model development process: \textbf{model selection} (choosing the more accurate model for a given input) and \textbf{counterfactual simulation} (figuring out a model's behavior on perturbed inputs). For each task, we conduct human studies to evaluate the effect of \textbf{saliency-based} explanations on humans' task performance. As the two tasks we study are ones typically encountered by machine learning engineers, we focus on users with machine learning experience.

To evaluate the effect of explanations for each task, we compare the accuracy of users not provided with any explanation to those provided with explanations. Additionally, we consider two different explanation access models: users learning from training examples with explanations then performing a downstream task without the explanations, and users with access to explanations during both training and performing the task. We evaluate SmoothGrad and GradCAM \citep{smilkov2017smoothgrad, selvaraju2017grad}, two popular saliency-based explanation methods. We additionally evaluate an ``oracle'' method, which is a synthetic explanation designed to be highly indicative of the correct answer for the tasks by highlighting according to simple rules and serves as an approximate upper bound of the extent to which saliency maps could help. We do not find sufficient evidence to conclude that saliency-based explanations, \emph{including our oracle method}, are useful for either of the tasks. In some cases, explanations enable participants to better describe the model's behavior, but still fail to improve their decisions about individual examples. In addition, despite the fact that oracle explanations are generated according to a simple set of rules, participants had widely varying interpretations of these explanations. Our findings suggest that significant additional work is needed to make saliency-based explanations truly ``interpretable'' to humans \citep{zhou2022exsum}. 

\begin{figure}
  \centering
  \includegraphics[width = \textwidth]{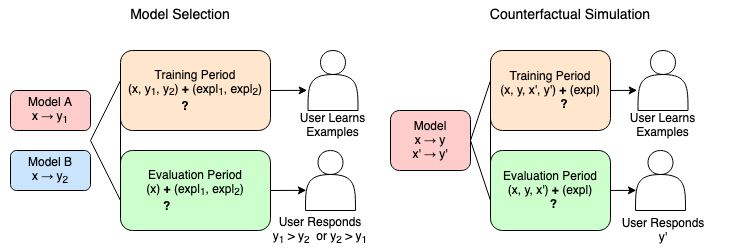}
  \caption{An overview of our user study tasks. Left: For model selection, we have two models, Model A and Model B, that map inputs $x$ to outputs $y_1$ and $y_2$ respectively. In training, the user has access to the inputs and outputs and potentially explanations. In evaluation, the user must determine based on the input and possibly explanations which model will more likely predict correctly. Right: For counterfactual simulation, we have a model that maps an input $x$ to output $y$ and a perturbed input $x'$ to output $y'$. In training, the user has access to both inputs $x, x'$ and both outputs $y, y'$ and potentially an explanation for $y$. In evaluation, the user must determine based on the inputs, $x, x'$ and possibly an explanation for $y$, whether the prediction for the perturbed image will differ from $y$.}
\end{figure}

\section{Related Work}

Determining whether an explanation helps a user understand a model better or more accurately can take on many forms. As a result various evaluation methods have been developed. Automated evaluations use a proxy metric to measure some property of an explanation. As a sanity check for what information explanations can encode, \cite{adebayo2018sanity} check whether explanations are sensitive to the training data used or the model parameters by randomizing each and comparing the resulting explanation with the original. Other work, including studies by \cite{samek2016evaluating} and \cite{hooker2019benchmark}, evaluates explanations by checking whether removing the most highlighted features results in a large change in the model's decisions. In a study by \cite{zhou2022feature}, explanation methods are evaluated on their ability to correctly highlight influential features. While these evaluations may determine whether explanations meet some criteria, it is also important to evaluate explanations on specific tasks and to study how humans interact with explanations. 

Other works propose task-specific and human-in-the-loop evaluations. \cite{adebayo2020debugging} introduce a method to evaluate an explanation's ability to diagnose model bugs, and \cite{bansal2021does} evaluate whether explanations help people perform better on tasks when working on a human-AI team. \cite{hase2020evaluating} and \cite{pruthi2022evaluating} perform evaluations of the extent to which an explanation can help a student model or a person simulate a teacher model on a test set after being exposed to a training set. \cite{lage2019evaluation} test whether explanations improve a human's ability to simulate model predictions or diagnose model issues. In a similar manner, \cite{chen2022use} present an automated method to test whether explanations can help an algorithmic agent on the downstream tasks of forward simulation, model debugging, and counterfactual reasoning. They also compared the results of the automated simulation to human performance on the corresponding tasks and found similar results, particularly that LIME provided a significant improvement. \cite{chu2020visual} test whether explanations for a model's predictions help on a task where people estimate the age of a person in an image given the model's prediction. \cite{schuff2022human} evaluate the factors affecting people's interpretations of explanations and provides adjustments for better calibrated understanding.

One limitation of some user study evaluations is that the tasks being evaluated do not reflect the needs of real-world developers and users of machine learning systems.
For example, in \cite{chen2022use}, the counterfactual prediction task is done on a synthetic problem with two variables. This makes it difficult to determine whether the findings will transfer to settings developers more regularly encounter, such as choosing the better model for deployment or determining when and why a model fails. In order to create a sense of practicality in our user study, the task scenarios are presented with context as to why we are interested in the particular task. For example, users were asked about how a model would behave on blurred images which may result from weather or equipment wear. 

We base aspects of our study design on several of the pieces of related work described above. One notable aspect of simulatability evaluations that \cite{pruthi2022evaluating} address is the possible leakage of information through explanations on the test set. As an extreme example, if for a bird detection task, the explanation always highlights the upper half of the image if the model detects the bird, and always highlights the lower half otherwise, this explanation would result in a perfect score if provided on the test set. In order to account for this information leakage, \cite{pruthi2022evaluating} only present explanations for examples in the training set, and we include a similar setup for comparison to the setting without explanations. Additionally, for accurate models, users may be able to perform the task themselves without considering the model's reasoning, and to account for this \cite{pruthi2022evaluating} and \cite{chu2020visual} balance the number of examples the user sees with and without model errors. We follow a similar idea of distinguishing the task from simulation. 

\section{Experiment Setup}

\subsection{Datasets}

We use the Visual Genome dataset \citep{krishna2017visual} to curate custom datasets for model training, because its images contain objects appearing in real-world contexts (to better simulate real-world scenarios) and because its annotations make it possible to automate creation of models with real-world failure modes like sensitivity to spurious features.
We use it to formulate binary classification tasks that require models to detect whether there is a \textbf{train} or \textbf{bus} in an image.

We create two datasets. The \textbf{baseline dataset} contains 3400 images, with balanced positive and negative labels. The positive class images all contain buses or trains, and the negative class images contain neither. The \textbf{mislabeled dataset} also contains 3400 images and is label-balanced, but the positive class is completely mislabeled. For this dataset, half of the images are in the positive class and contain images of cars, vans, or trucks and excludes images that contain buses of trains according to the image annotations. The other half are in the negative class and are images that contain no vehicles at all. As a result, there is no overlap between the positive classes of the baseline and mislabeled dataset while the negative classes may overlap. This dataset represents a mislabeling that results from choosing images for the positive class from a similar or broader class of images. We discuss additional criteria for selecting images to prevent user confusion in Appendix \ref{app-img}.

We choose to focus on a simple task along with an extreme case of mislabeling so that the resulting model behaviors are more likely to be conveyed by saliency maps. If a saliency map fails to provide useful information on this task, it is unlikely that it would be helpful for understanding models performing the more complex tasks to which machine learning methods are often applied.

\subsection{Tasks}

\paragraph{Model Selection} The model selection task is motivated by the real-life scenario of having to identify the better of two models to deploy. While these decisions are often made by inspecting the models' predictions, explanations can in principle provide additional signals, resulting in potentially better decision-making. In our study, users are presented with the task of choosing the better of two models both designed for the task of detecting trains and buses. We provide the additional context of a company wanting to understand which model would be better to deploy in different settings. 

Specifically, the task consists of 3 questions asking which model they would choose to deploy at a bus/train station if the company prioritizes minimizing false positives/negatives; along with 12 questions asking which model would be more likely to correctly classify individual images. The questions about individual images provide a finer-grained indication of the extent to which the person understands the difference in model behavior. We also include 32 training examples that provide the user with example images and each model's prediction and confidence. The training examples and 12 questions on individual images are evenly balanced between images that contains buses or trains and images that do not. Half of the negative images contain cars without trains. We train the two models by fine-tuning a pretrained ResNet18 on the baseline dataset and on the mislabeled dataset. 

\paragraph{Counterfactual Simulation} The counterfactual simulation task is motivated by the real-life scenario of predicting how a model will behave when the input distribution shifts under some perturbation such as blurring or a change in lighting. It can be difficult to have access to data under these shifts, and explanations could help reason about model behavior in these situations. In our study, users are presented with the task of predicting whether a single model will change its prediction after an image is blurred. At the beginning of the study, we provide the additional context that a company wants to understand model behaviors under image blurring due to weather or camera wear. 

The tasks consists of 12 questions in which the user is presented with a pair of images that are the same except one has a blurred region. The blurred region has a width and length that is 60\% of the image's width and length respectively and is randomly placed according to a uniform distribution over the image. The user is given the model's prediction on the unblurred image and asked whether the prediction will change for the blurred image. Similarly to model selection, we include 32 training examples that provide the user with example pairs of images and the model's prediction on both. The training examples are evenly balanced between images that contains buses or trains and images that do not. For both the evaluation questions and the training examples, half of the image pairs have a positive label and the other half have negative labels. Half of negative examples include a car or other non-train/bus vehicle. We use a pretrained ResNet18 fine-tuned on the baseline dataset for this task.

\paragraph{Qualitative Questions} At the end of both tasks, we ask the user to describe their understanding of how models work and what parts of images and explanations helped them determine their answers. 

\subsection{Explanation Methods}

We focus on explanations that are in the form of saliency maps. 
We focused our  study on the SmoothGrad \citep{smilkov2017smoothgrad} and GradCAM \citep{selvaraju2017grad} methods as they resulted in the simplest maps while also focusing on vehicle-related objects, and example visualizations are provided in Appendix \ref{app-a}. Other methods that we considered were gradient-based explanations \citep{simonyan2013deep}, Gradient$\odot$Input \citep{shrikumar2016not}, Integrated Gradients \citep{sundararajan2017axiomatic}, LIME \citep{ribeiro2016should} and RISE \citep{petsiuk2018rise}.
The simplicity criteria that resulted in the choice of SmoothGrad and GradCam were the number of different highlighted regions and the noisiness of explanations.

\begin{figure}
  \centering
  \includegraphics[width=0.46 \textwidth]{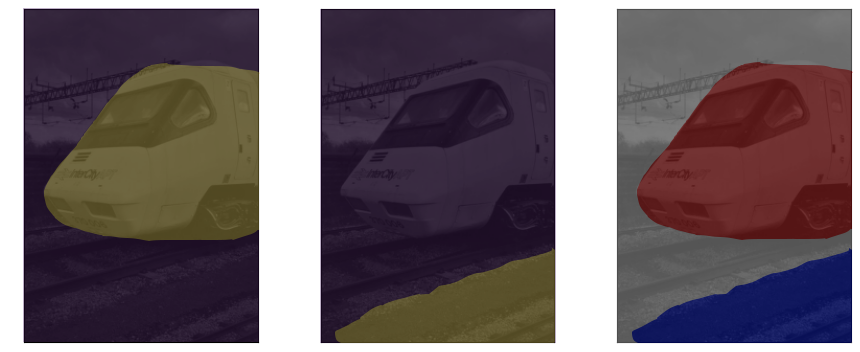}
  \hfill
  \includegraphics[width=0.28 \textwidth]{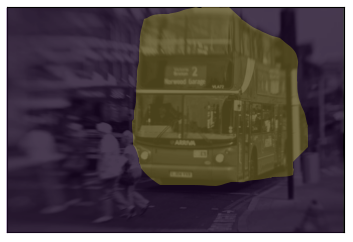}
  \includegraphics[width=0.145 \textwidth]{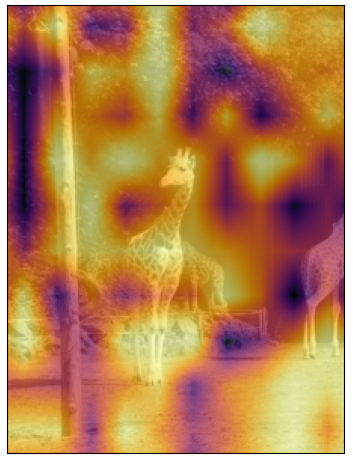}
  \caption{Examples of the oracle method on each task. The left set of three images is an example of the explanation for model selection where the leftmost image corresponds to a model detecting the train, the middle image corresponds to a model not detecting the train, and the rightmost image corresponds to the difference in the two attributions. The right set of images shows the oracle method for counterfactual simulation where the left image shows the explanation for a model detecting a bus and the right image shows the explanation for a model not detecting a train or bus.}
  \label{fig2}
\end{figure}

In addition, we consider a synthetic oracle explanation that directly reveals the correct user behavior. This is an example of an explanation that ``leaks'' information on the evaluation set as discussed in \citet{pruthi2022evaluating}. Assuming that saliency methods in practice do not reveal as much information as our oracle, this oracle explanation provides an approximate upper bound on the benefit of saliency maps when provided for the training and evaluation set. Additionally, we can study what conclusions users form when provided with the oracle explanation only for the training set. At a high level, the oracle explanation is designed such that most of the evaluation images with the same answer have the same highlighting pattern. More details are provided in Appendix \ref{app-b}.

\section{Results}


\paragraph{Quantitative Results}

We evaluated the statistical significance of the differences in results between the no explanation setting and each of the six other settings. The null hypothesis is that the average accuracy for the two settings is the same. The results of these comparisons (with $p$-values) can be seen in Table 2. 

As shown in Table \ref{acc-table}, we fail to reject the null hypothesis of explanations having no effect for tasks and explanation methods, including oracle explanations.
We consider a result significant at $p = 0.0083$ after applying a Bonferroni correction for multiple comparisons. 
We provide bar chart visualizations of these results in Appendix \ref{app-a}. These results suggest that, even on a simple version of a realistic task, there is a lack of evidence to support the conclusion that explanations contribute to user performance.

\begin{table}[ht]
  \caption{Average accuracy with 95\% confidence intervals on the two tasks across settings}
  \label{acc-table}
  \centering
  \begin{tabular}{lcccc}
    \toprule
      & \multicolumn{2}{c}{Model Selection} & \multicolumn{2}{c}{Counterfactual Simulation} \\
      & Accuracy & $p$-value & Accuracy & $p$-value \\
    \midrule
    No Explanation & 67.8\% $\pm$ 9.71\% & & 56.9\% $\pm$ 11.5\% &   \\
    \midrule
    SmoothGrad (Train only) & 80.0\% $\pm$ 9.11\% & 0.074 & 65.3\% $\pm$ 12.3\% & 0.459   \\
    GradCAM (Train only) & 68.6\% $\pm$ 8.92\% & 0.906 & 63.1\%  $\pm$ 10.4\% & 0.438 \\
    Oracle (Train only) & 72.2\% $\pm$ 9.31\% & 0.518 & 54.2\% $\pm$ 11.6\% & 0.740   \\
    \midrule
    SmoothGrad (Train and Eval) & 48.9\% $\pm$ 10.4\% & 0.010 & 63.3\% $\pm$ 11.2\% & 0.398  \\
    GradCAM (Train and Eval) & 66.7\% $\pm$ 9.79\% & 0.875 & 58.3\% $\pm$ 11.5\% & 0.867  \\
    Oracle (Train and Eval) & 75.6\% $\pm$ 8.93\% & 0.249 & 68.1\% $\pm$ 10.8\% & 0.171  \\
    \bottomrule
  \end{tabular}
\end{table}


\paragraph{Qualitative Descriptions}

In order to understand how explanations influenced users' understanding models, we asked users to ``describe what parts of the images/explanations contributed to your answers'' as well as to describe how they thought the models differed for model selection or how they thought the model made its prediction for counterfactual simulation. The full results for these questions are provided in Appendix \ref{app-c}. 

\begin{table}[ht]
  \caption{Percent of user descriptions that mentioned a significant aspect of model behavior}
  \label{user-table}
  \centering
  \begin{tabular}{llll}
    \toprule
      & No Explanations & Non-oracle & Oracle\\
    \midrule
    Model Selection: & & & \\
    Described false positive rate on cars/vehicles & 16.7\% & 12.5\% & 41.7\%   \\
    \midrule
    Counterfactual Simulation: & & & \\
    Described no change in negative predictions & 83.3\% & 12.5\% & 33.3\%   \\
    \bottomrule
  \end{tabular}
\end{table}

For model selection, the largest difference in how the models behave is the fact that one model acts as a car detector while the other acts as a bus and train detector. As seen in Table 2, the portion of users that mentioned this behavior is much larger for those provided with the oracle explanation. Based on this, it seems the design of the oracle explanation helped bring attention to this behavior in particular. However, despite this difference, the oracle did not significantly improve the task performance. This may be because while those provided with the oracle explanation more often described the behavior on cars, others were also aware of it but did not see it as a key component of the model. 

For counterfactual simulation, the qualitative descriptions of models provides some evidence that explanations can cause people to miss some of the simpler patterns. As seen in Table 2, the fraction of users that described the lack of change in blurred images that originally had negative images is lower for users with non-oracle explanations. In particular, some users provided with the GradCAM explanation gave descriptions such as ``The main problem is that if a vehicle is at an angle that's outside of a 45 degree view'', or ``It looks for streaks of similar colour that form a pattern''.

\begin{table}[ht]
  \caption{Percent of users with different usage of explanations}
  \label{usage-table}
  \centering
  \begin{tabular}{lll}
    \toprule
      & Only Training & Training and Evaluation\\
    \midrule
    Mentioned matching to similar examples & 38.9\% & 5.55\%   \\
    \midrule
    Only described reading explanations & 0\% & 16.7\%   \\
    \bottomrule
  \end{tabular}
\end{table}

The use of explanations varied largely between the settings where they were only provided for training examples and the settings where they were provide for all images. The majority of users provided with explanations only for training examples mentioned either basing their answer on similar examples or basing their answer on general patterns across images. Users provided with explanations for all images primarily describe how they read the explanation for each image and concluded an answer from that. This emphasis on reading individual examples may be one of the reasons as to why people may not have noticed some of the simpler patterns on the counterfactual simulation task when provided explanations on all images and why the oracle explanation did not perform as well as expected. 

\section{Discussion}

We have provided an evaluation of saliency maps based on their ability to help users familiar with machine learning perform model selection and counterfactual simulation. While we did not find evidence of improvement, one major limitation of our study is that its small sample size does not allow us to detect very small improvements in accuracy resulting from explanations in the presence of large variability across human participants. 
Other limitations include the number of examples users had access to---it is still possible that users might perform better with either more examples or additional modes of explaining. In particular, the varying descriptions of how the oracle explanations was used despite its simple rules for generations suggest that in order for people to make stronger conclusions, more examples are needed. Understanding the relationship between the trust users have in the explanations, what kinds of conclusions users draw, and the number of examples provided is one potential future direction of research. 

In addition to the mislabeling setting we study in our evaluation, there are other real-world settings that we have not considered such as spurious correlations in data, verifying a model's diagnosis, or a distributional shift at deployment. Evaluating whether feature attribution methods help people perform these tasks remains an open area of study. On top of different settings, considering other explanation methods such as concept-based explanations is another possible direction for future work. There remains many unexplored aspects of current explanation methods, and our findings suggest that when evaluating these methods, we should carefully consider what makes an explanation truly ``interpretable'', and how different users interpret an explanation.

\bibliographystyle{plainnat}
\bibliography{neurips_2023}

\newpage
\appendix

\section{Additional Experiment Visualizations}\label{app-a}

\begin{figure}[ht]
  \centering
  \includegraphics[width=\textwidth]{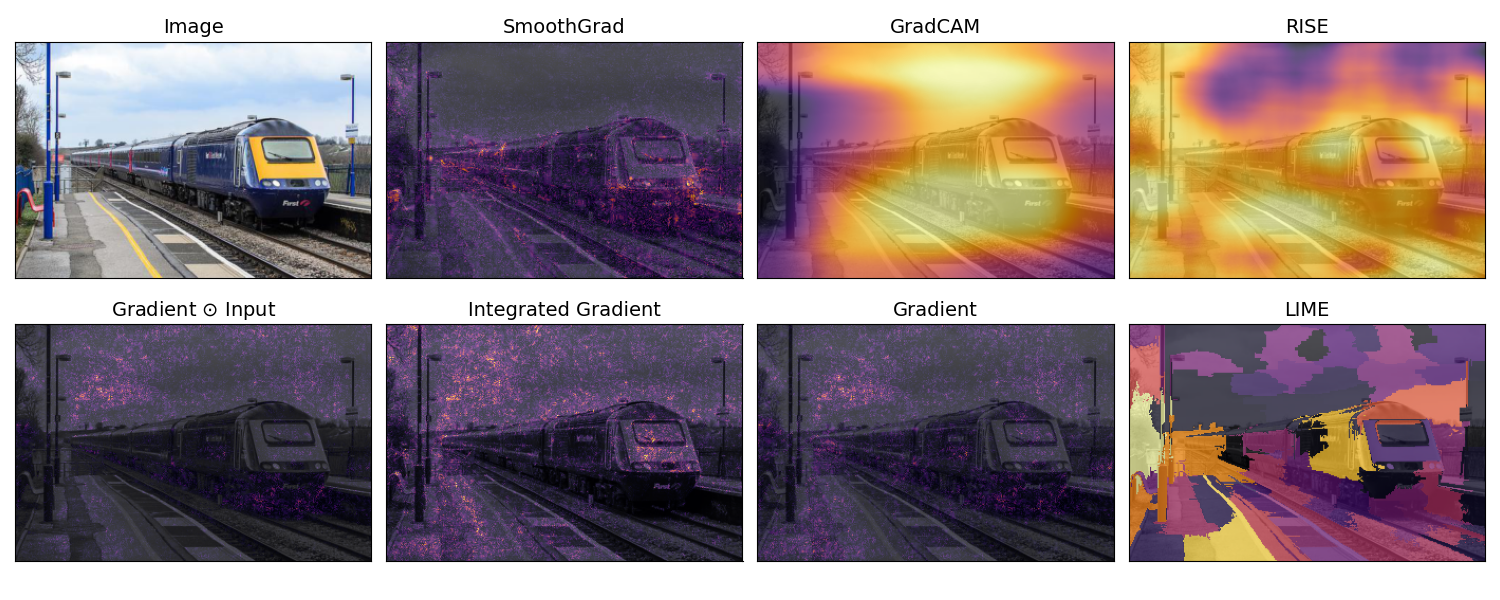}
  \caption{Example of explanation methods for an image of a train}
  \label{fig4}
\end{figure}

\begin{figure}[ht]
  \centering
  \includegraphics[width=\textwidth]{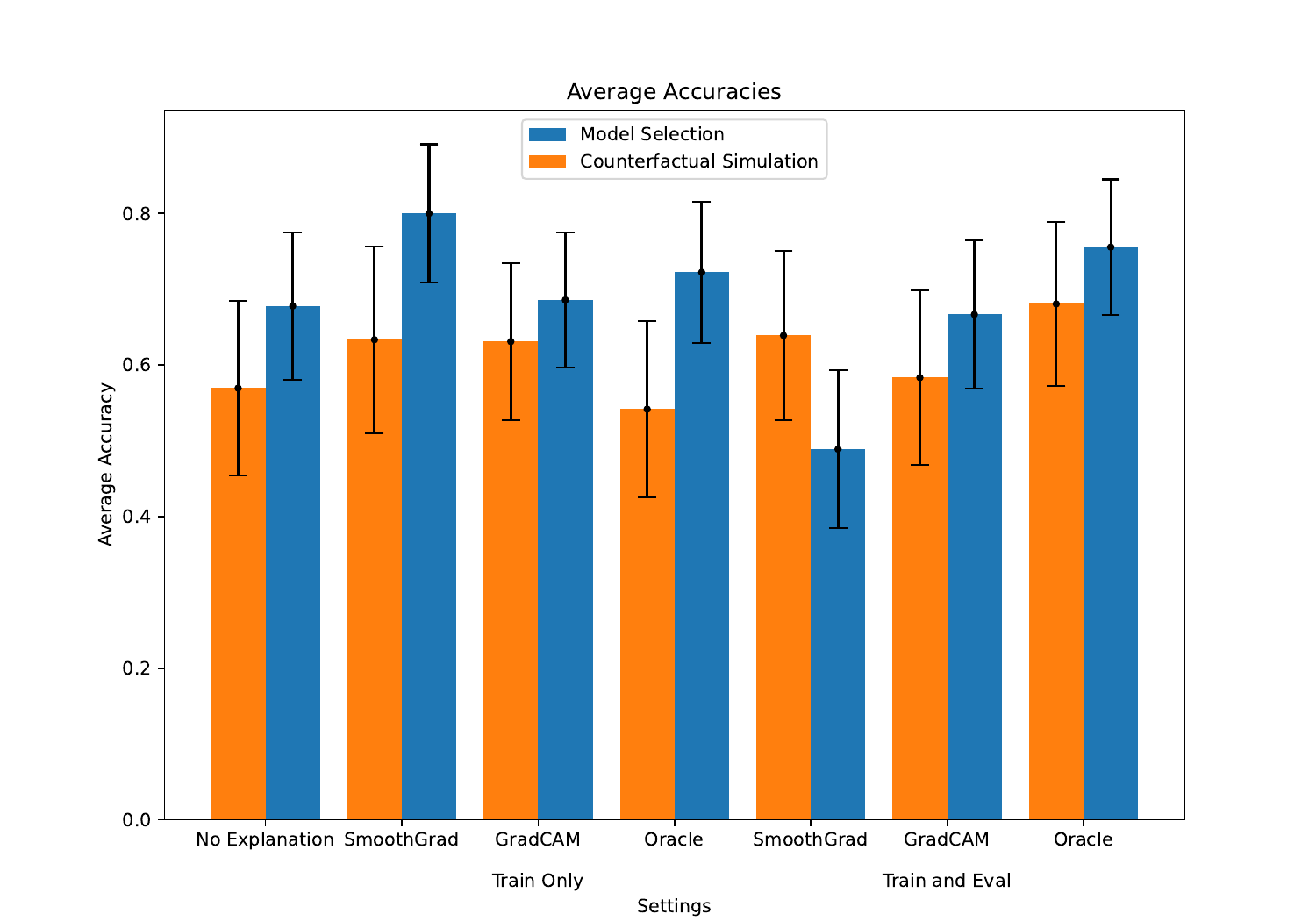}
  \caption{Bar chart of average user performance on model selection and counterfactual simulation across different settings with a 95\% confidence interval}
  \label{fig5}
\end{figure}

\clearpage
\section{Image Selection Criteria} \label{app-img}
In order to minimize confusion to the user, we have additional criteria for the images in the datasets. The first is that if the image is labeled as containing an object, the dataset's bounding box for that object must have dimensions that are at least 30\% of the image's corresponding dimension, to ensure clear visibility. The second is that if the image is not supposed to contain some object, it must not be in the image's labeled objects nor be mentioned in the image's caption. This is to minimize unintentional mislabelings in the negative class as sometimes an object appears in the image and is not labeled, and checking the caption acts an additional filter. 

\section{Oracle Explanation Generation} \label{app-b}

We provide in detail the rules used to generate the orcale explanation. 

For model selection, the oracle explanation was generated according to three rules:
\begin{enumerate}[leftmargin=*]
    \item If the model detects a bus or train in the image, highlight a vehicle (buses or trains are prioritized) in the image if present, and highlight any rectangular region of the image.
    \item If the model does not detect a bus or train, highlight some other object in the image. 
    \item If the two models make the same prediction, they highlight the same object but the one with higher confidence highlights the object brighter. 
\end{enumerate}
This explanation conveys the exact answer for the model selection task as the difference in highlighting will always indicate the correct answer. For counterfactual simulation, the explanation was generated according to the following rules:
\begin{enumerate}[leftmargin=*]
    \item If the model detects a bus or train in the image, highlight a vehicle (buses or trains are prioritized) in the image if present, and highlight a rectangular shape otherwise.
    \item If the model does not detect a bus or train, generate a randomly highlighted map. 
\end{enumerate}
The oracle explanation is mainly designed to convey that when the model does detect a bus or train, if most of the object is blurred the prediction could change and when the model does not detect a bus or training, no matter what is blurred, the prediction does not change. To express that in a saliency map, positive predictions highlight a vehicle when present and negative predictions have a uniformly scattered map. We choose not to highlight some other object for negative predictions as this could indicate that the absence of the object may lead to a change in prediction. This explanation does not have a direct correlation in highlighting with the correct answer for all questions for the counterfactual simulation task but does provide a high-level picture of the model that could improve a user's performance. 

\section{Additional Experimental Details}
\subsection{Datasets}

Each dataset used is split into a training and validation set with 3000 of the images belonging to the training set. We create an additional test set of 800 images which we sample from to use for the tasks in our study: 200 contain a bus or train but no cars, vans, or trucks, 200 contain a car, van, or truck but no buses or trains, and 400 contain none of the vehicles. 

\subsection{Participants}

We recruited 43 participants with some ML experience, and asked each person to perform both tasks. Each user was randomly assigned to one of seven settings: no explanation, or $\{$SmoothGrad, GradCAM, oracle$\} \times \{$training time access only, training and evaluation time access$\}$. To ensure that participants understood the task and were providing proper responses, participants were required to answer attention check questions. We excluded one participant's response due to a misunderstanding of directions. We paid all users \$20 per hour. 

\clearpage
\section{User Interface}\label{app-d}
\subsection{Model Selection}
\begin{figure}[ht]
  \centering
  \includegraphics[width=\textwidth]{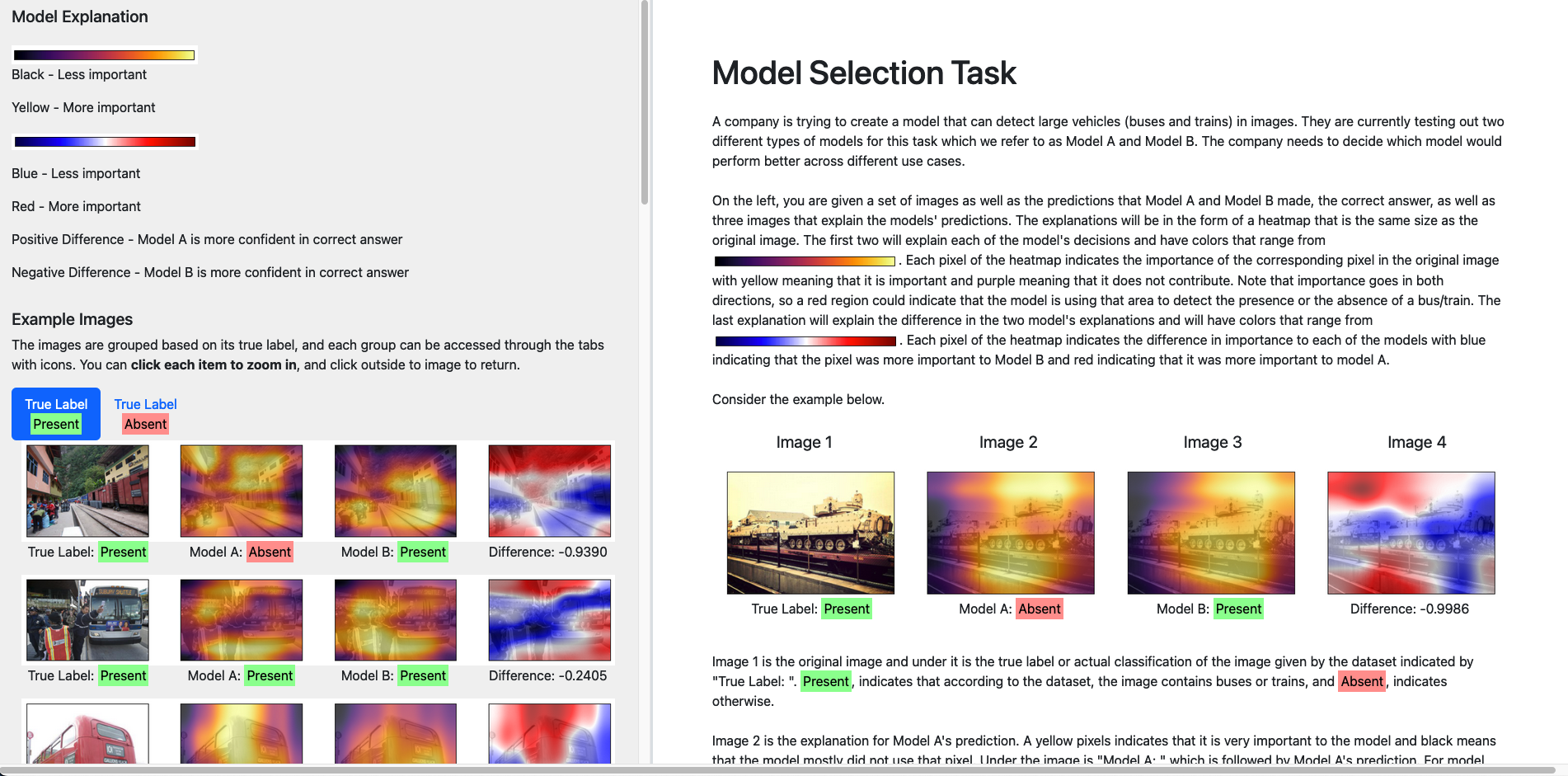}
  \caption{Screenshot of model selection user interface}
\end{figure}

\begin{figure}[ht]
  \centering
  \includegraphics[width=\textwidth]{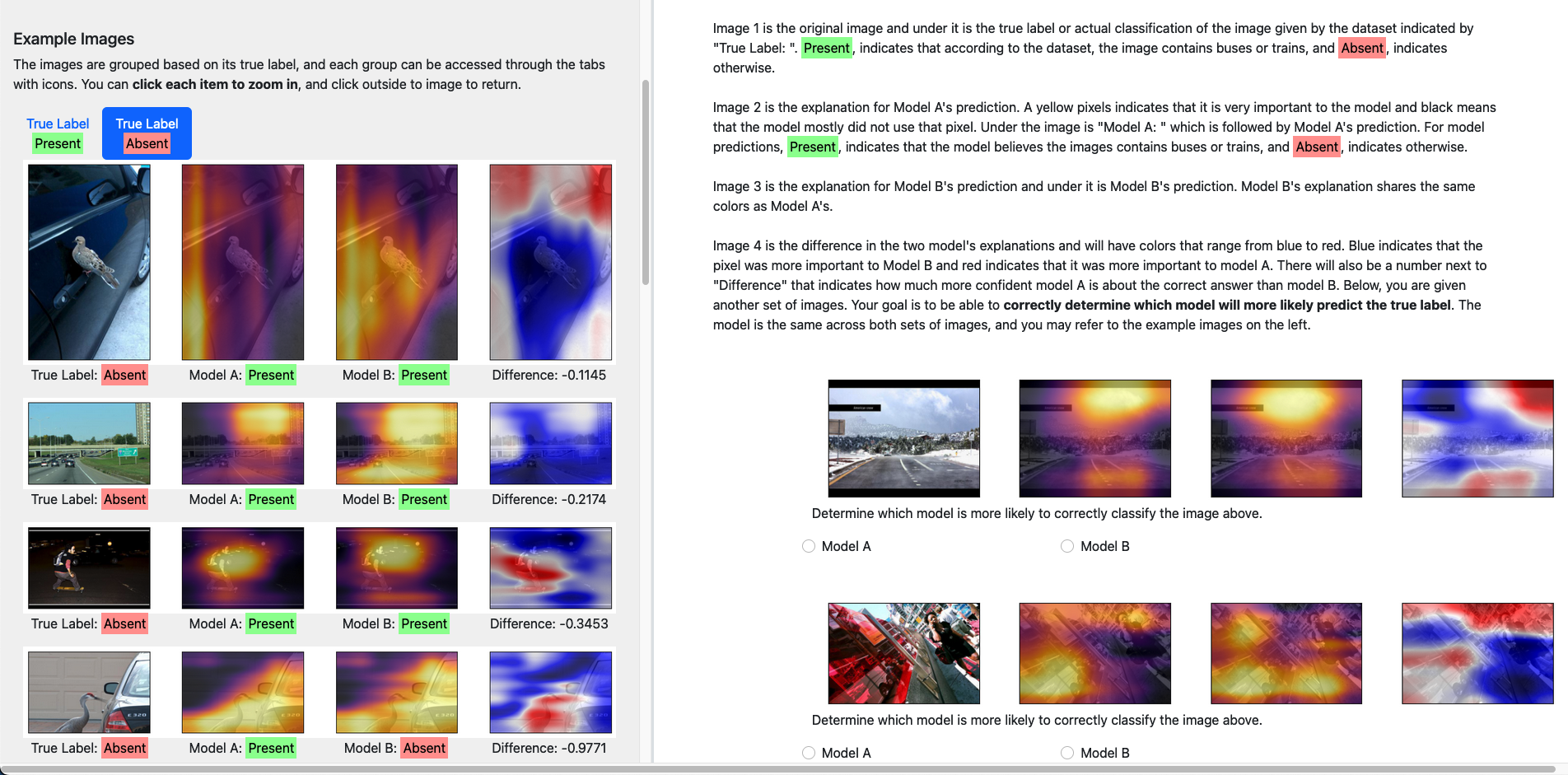}
  \caption{Screenshot of model selection user interface}
\end{figure}

\begin{figure}[ht]
  \centering
  \includegraphics[width=\textwidth]{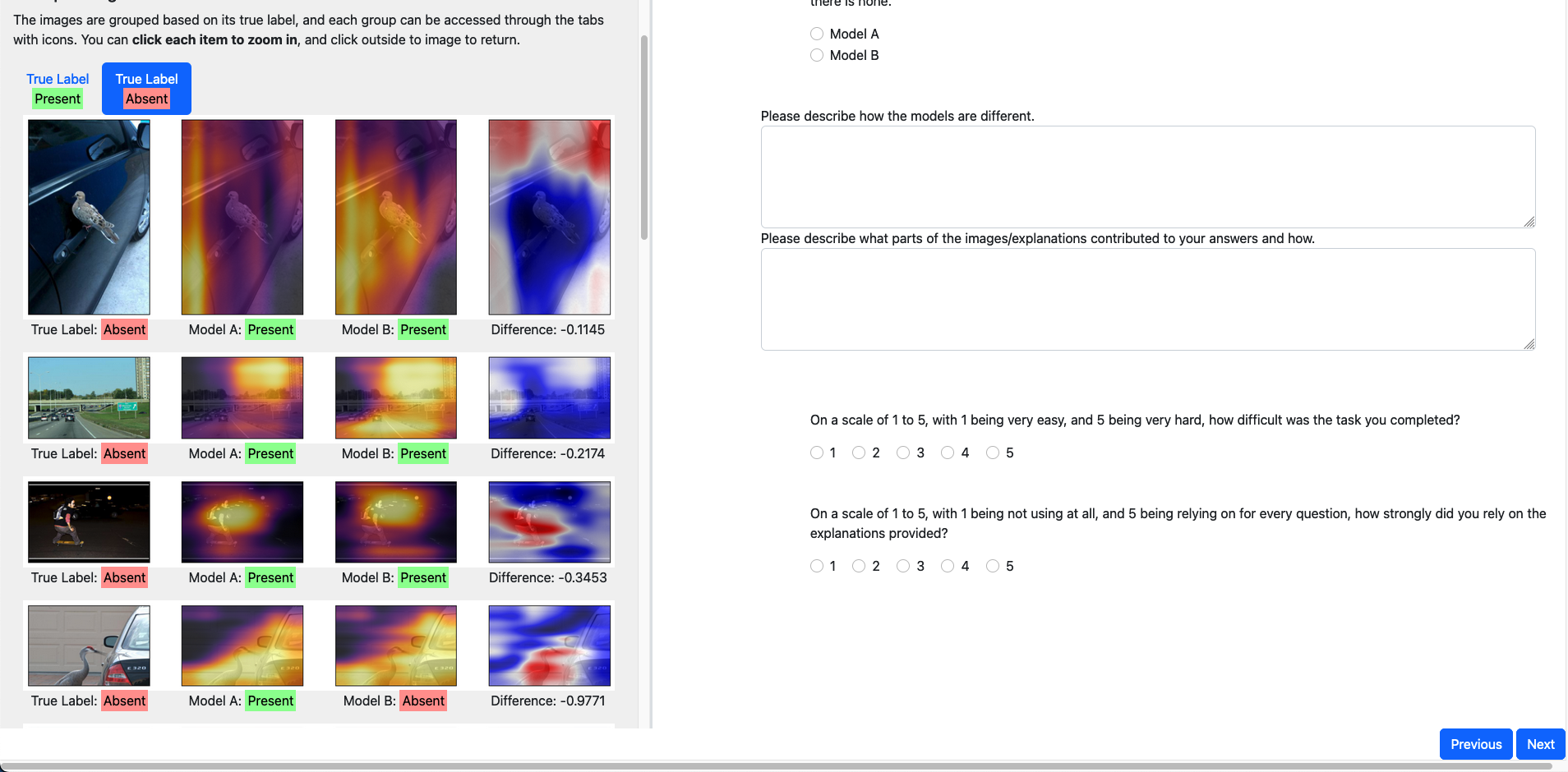}
  \caption{Screenshot of model selection user interface}
\end{figure}

\clearpage
\subsection{Counterfactual Simulation}
\begin{figure}[ht]
  \centering
  \includegraphics[width=\textwidth]{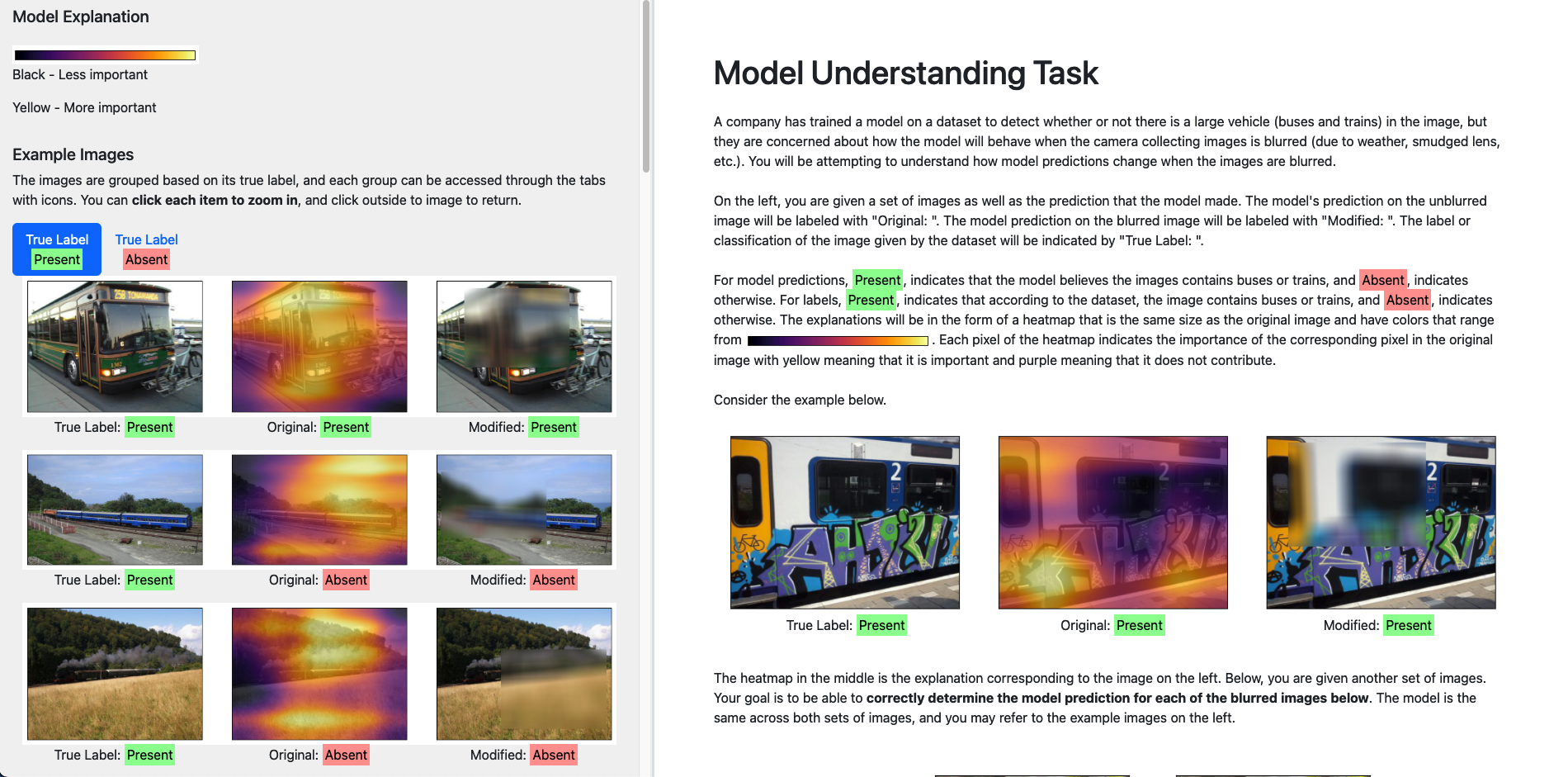}
  \caption{Screenshot of counterfactual simulation user interface}
\end{figure}

\begin{figure}[ht]
  \centering
  \includegraphics[width=\textwidth]{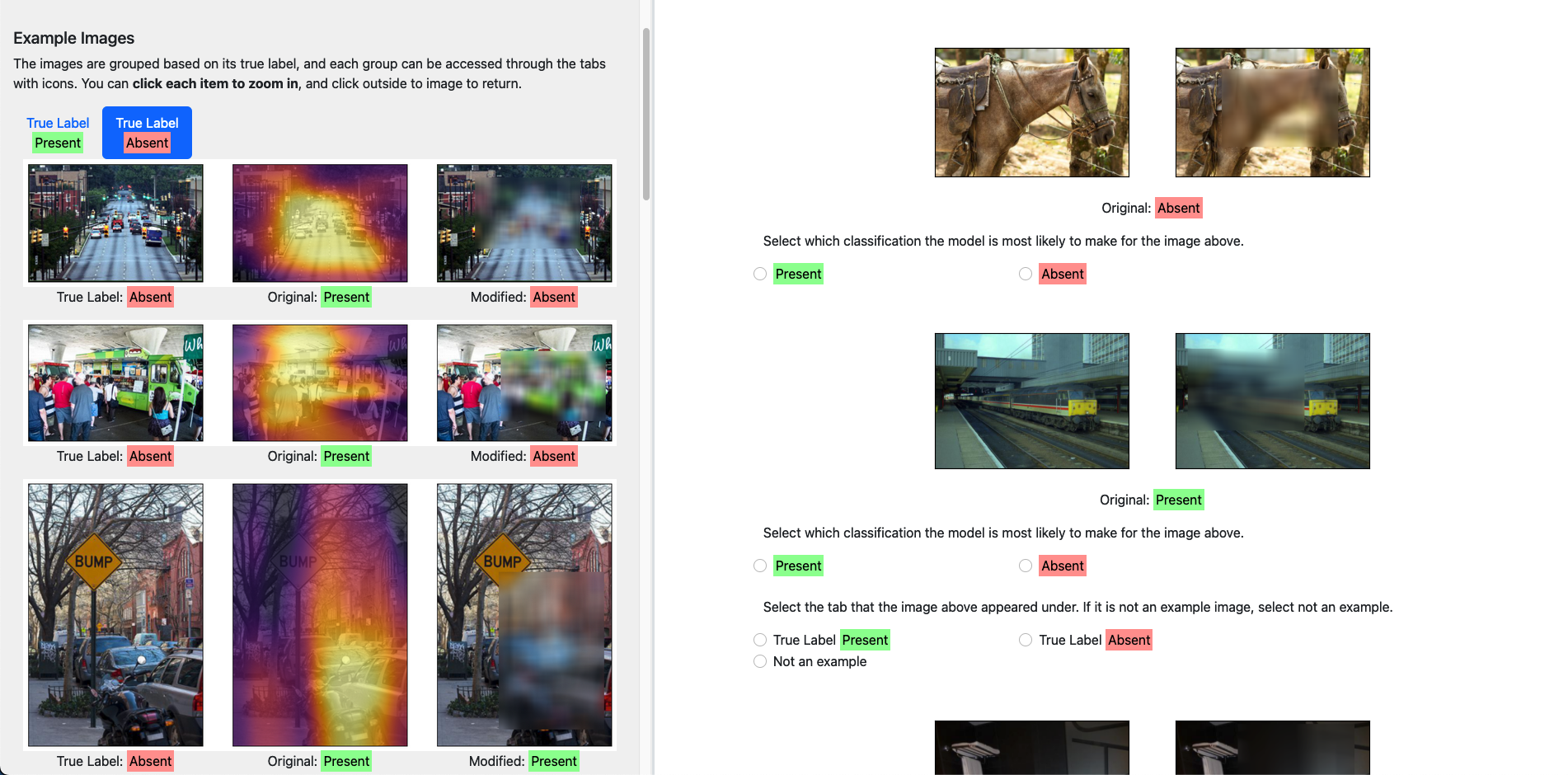}
  \caption{Screenshot of counterfactual simulation user interface}
\end{figure}

\begin{figure}[ht]
  \centering
  \includegraphics[width=\textwidth]{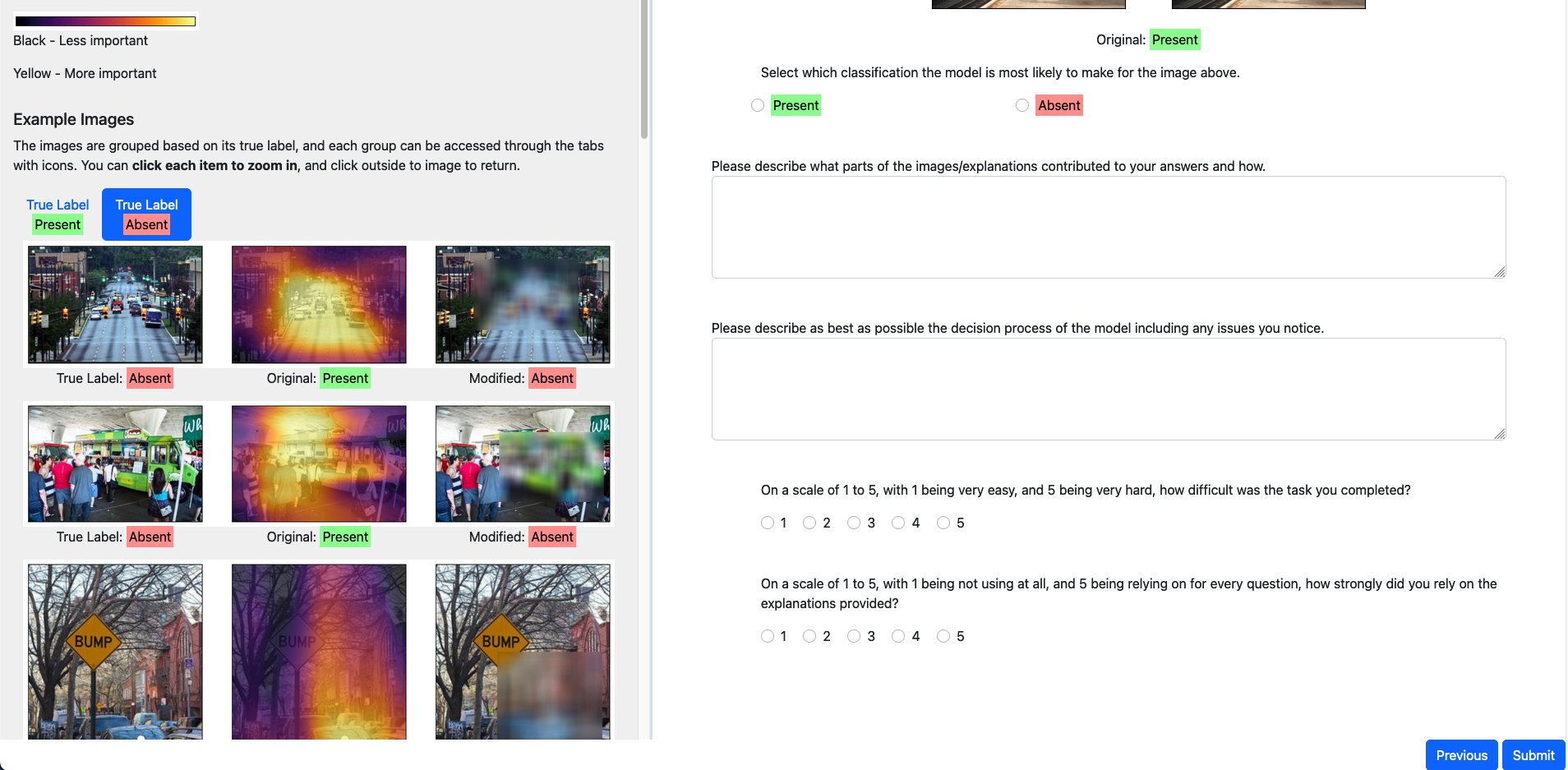}
  \caption{Screenshot of counterfactual simulation user interface}
\end{figure}

\clearpage
\section{Full Qualitative Results}\label{app-c}

\subsection{Model Selection}

\begin{table}[ht]
  \caption{Responses to ``Please describe how the models are different. '' for Model Selection for the no explanation setting}
  
  \centering
  \begin{tabular}{p{0.95\linewidth}}
    \toprule
    Response (\textit{No Explanation})\\
    \midrule
    Model A sometimes says there isn't a train when there is one, but always correctly identifies when there isn't one. \\
    \\
    Overall, Model B's performance is much more consistent. It has a Type 1 error rate (false positive rate) of 3/8, and a Type 2 error (false negative rate) of 1/16, meaning that it rarely will ``miss'' a train or bus. On the other hand, Model A does not perform consistently at predicting positives or negatives. It has a Type 1 error rate of 1/2, and a Type 2 error rate of 7/16. Having an error rate close to 1/2 when you only have 2 options is pretty rough :p Model B tends to predict images that contain buildings or people incorrectly, and incorrectly predicts a false positive for 3/7 of images centrally containing people.	\\
    \\
    Model A seems to be dependent on the presence of people and possible faces in particular to state whether or not a train is present. It also seems to do more poorly with trains in the background/when it it is not a focus point of the image.	\\ \\ 
    Model A is better at predicting whether an image has a ``large'' vehicle if it has another smaller vehicle to compare it to. Model B seems to default to predicting ``Present'' whereas Model A seems to default to predicting ``Absent''	\\ \\
    Model B is more likely to correctly classify the true label, if the true label is ``present''. Model A is more likely to correctly classify the true label, if the true label is ``absent''.	\\ \\
    - B is pretty good; it can catch vehicles in reflections and parse buses and trains in most situations - A is good at detecting when there are only humans in an image \\
    \bottomrule
  \end{tabular}
\end{table}

\begin{table}[ht]
  \caption{Responses to ``Please describe how the models are different. '' for Model Selection for the GradCAM (Training only) setting}
  
  \centering
  \begin{tabular}{p{0.95\linewidth}}
    \toprule
    Response (\textit{GradCAM, Training only})\\
    \midrule
    Model A uses more input from the environment than does Model B; Model B focuses more on the objects in the image.	\\ \\
    B focuses more on what is in the middle of the image while A focuses around the edges. Effectively, B hyperfixates on specific points, which causes it to not look at the big picture.	\\ \\
    Model A is less likely to predict trains/buses but better with people; however, it still does on occasion.	\\ \\
    model A seems to do better at detecting vehicles even with the presence of \\ animals; model B seems to more freely classify images as containing a vehicle even when it doesn't	\\ \\
    It appears that the difference between the models is that A tends to over-label images as ``Present'' even if there is no train / bus. B tends to be more accurate with what the labels, but overall tends to label things as more ``Absent'' than not. The ``type'' of errors between A and B are opposite where A over-labels ``Present'' and B over-labels ``Absent''.	\\ \\
    B seems to be more accurate and also more confident than A (at least when the bus/train was present)	\\ \\
    Model A appears to be less confident than Model B across both present and absent examples, maybe except where there are non-bus/train things taking up most of the frame, especially people. Model B seems to place more emphasis on vehicle-like objects in the scene, while Model A seems to often pick up background. \\
    \bottomrule
  \end{tabular}
\end{table}

\begin{table}[ht]
  \caption{Responses to ``Please describe how the models are different. '' for Model Selection for the SmoothGrad (Training only) setting}
  
  \centering
  \begin{tabular}{p{0.95\linewidth}}
    \toprule
    Response (\textit{SmoothGrad, Training only})\\
    \midrule
    Model B is much more likely to detect trains when the train is present. The two models have similar performance when detecting trains when the train is not present. 	\\ \\
    Model A looks at the surrounding of the image compared to Model B. Model A is more likely to determine that the train/bus is not in the image if the object is not front and center.	\\ \\
    Model A is less sensitive to false positives on people, but more sensitive to false positives with cars. Model B works generally better for the least false negatives, but has more false positives when people are in the image. However, it is better than Model A with respect to false positives for cars.	\\ \\
    Model B seems to rely more upon shape image classification of large vehicles, whereas Model A has a harder time identifying vehicles when there is additional noise in the image.	\\ \\
    It seems like model B does better overall, though it struggles when there are individual subjects, especially people in the center as it'll return present, but it does well with individual points. Model A struggles on accuracy, particularly when there are a lot of things in the photo, as it will declare present then, but absent when there are less items.  \\
    \bottomrule
  \end{tabular}
\end{table}

\begin{table}[ht]
  \caption{Responses to ``Please describe how the models are different. '' for Model Selection for the Oracle (Training only) setting}
  
  \centering
  \begin{tabular}{p{0.95\linewidth}}
    \toprule
    Response (\textit{Oracle, Training only})\\
    \midrule
    I think the models have different focuses like people, cars, and other vehicles. \\ \\
    Model A commonly mistakes cars for buses or trains. Model A also frequently misses buses/trains that are in the background. Model B is typically better at identifying trains that are present. \\ \\
    It seems like model B is better at predicting when trains appear, given that it looks for long horizontal smears and classify images as present when the smears are found. Model A seems to look for when there are smaller or larger blobs, and classify images as present when larger blobs are identified. \\ \\
    The models are fairly similar. On the whole Model B does better than Model A with people in them. \\ \\
    Model A seems to detect the main focus of the image, while Model B will detect things in the background, which can be either good or bad depending on where the vehicle is. Overall, Model B seems to be more confident than Model A, and Model A gets confused between large vehicles and cars. \\ \\
    Model A does a good job of penalizing non-vehicle pictures, but has a hard time telling the difference between cars and trains/subway. Model B does a good job of finding trains in inconspicuous places, but doesn't do a great job at identifying specific parts of a train (like the train head). It also has a lot of irrelevant false positives. \\ 
    \bottomrule
  \end{tabular}
\end{table}

\begin{table}[ht]
  \caption{Responses to ``Please describe how the models are different. '' for Model Selection for the GradCAM (Training and Evaluation) setting}
  
  \centering
  \begin{tabular}{p{0.95\linewidth}}
    \toprule
    Response (\textit{GradCAM, Training and Evaluation})\\
    \midrule
    model A finds all cars, not just bus and trains, and is sometimes not goo at detecting trains. Model B is good at buses and cars but also finds people. \\ \\
    B focuses more on trains while A focuses more on background \\ \\
    It is my understanding that Model B prioritizes varying colors, objects, etc, and filters accordingly whereas Model A prioritizes consistent fields (or the opposite of B) \\ \\
    Model A gives greater importance to space outside of the central object, while Model B focuses on parts of the train/car. \\ \\
    Model B is better at finding vehicles and marking them as important. Model A seems to focus on the background more. \\ \\
    Model A seems to rely on vehicle features like large front windows, or the head of a train/bus, to predict the existence of large vehicles, while model B seems to rely more on a series of windows or recurring rectangular patterns. \\
    \bottomrule
  \end{tabular}
\end{table}

\begin{table}[ht]
  \caption{Responses to ``Please describe how the models are different. '' for Model Selection for the SmoothGrad (Training and Evaluation) setting}
  
  \centering
  \begin{tabular}{p{0.95\linewidth}}
    \toprule
    Response (\textit{SmoothGrad, Training and Evaluation})\\
    \midrule
    B usually deems less pixels are important while A seems more pixels important. \\ \\
    model a seems to lean on the side of predicting that there is no train \\ \\
    The heatmap for B shows that Model B seems to place more importance on a greater number of aspects/points, whereas Model A is more focused. \\ \\
    Model A seems to be searching and prioritizing common vehicle features like wheels windows etc. It is good at seeing forms, but cannot easily differentiate between cars and large vehicles. Model B on the other hand is good at seeing figures that provide big contrast in the image, it is good at identifying between cars and buses for example but fails to identify completely different images without vehicles. \\ \\
    Model A tended to be overconfident (more false positives), and Model B tended to be more cautious (more false negatives). \\ \\
    Model A correctly guesses present more often when there are people in the image too, while model B seems to not be as affected, model B is also more confident in guessing for present. When the image is absent, model B seems to be less confident in the prediction vs model A when both predict absent. But model A seems to incorrectly predict present in absent situations, and model A has a low confidence in those situations. \\
    \bottomrule
  \end{tabular}
\end{table}

\begin{table}[ht]
  \caption{Responses to ``Please describe how the models are different. '' for Model Selection for the Oracle (Training and Evaluation) setting}
  
  \centering
  \begin{tabular}{p{0.95\linewidth}}
    \toprule
    Response (\textit{Oracle, Training and Evaluation})\\
    \midrule
    For the images without cars, Model B is more likely to assign a label of True even when there is no car, and Model A is more likely to assign a False label, even when there is a car. For the images with cars, Model B seems to be better at both detecting trains and ignoring cars. \\ \\
    Model B gets it right more often of the time but also comes up more in the difference. \\ \\
    Model B seems to more accurate than Model A. It has a lot more true positives and true negatives. Model A tends to miss trains/buses by focusing on the tracks, buildings, or people (as an indicator that there's no trains even when there's a train next to it). Model A seems to be detecting cars and saying ``Present'' because of it (as in the label is ``Present'' but it is focusing on a car and there's no train/bus present), whereas in those cases model B would say correctly say ``Absent''. However, model B has the issue of seeing a vaguely rectangle shape and say falsely say ``Present''. \\ \\
    They seem to be mainly similar, but Model A tends to use things that have smooth lines whereas Model B uses a combination of things with smooth lines and things with odd shapes \\ \\
    Model A looks for any coherent object in the image, whereas Model B always looks for a train or bus whether or not it is present.  \\ \\
    Model A is just a bad model. Model B actually gets it right most of the time. Maybe this means that model A pays more attention to background info. \\
    \bottomrule
  \end{tabular}
\end{table}

\begin{table}[ht]
  \caption{Responses to ``Please describe what parts of the images/explanations contributed to your answers and how. '' for Model Selection for the no explanation setting}
  
  \centering
  \begin{tabular}{p{0.95\linewidth}}
    \toprule
    Response (\textit{No Explanation})\\
    \midrule
    The lack of/appearance of trains/buses and which model got it correct \\ \\
    When making my decisions, I paid attention to the error rates for positive and negative images for both models, and when answering questions relating to which model would perform stronger, I looked at rough 'categories' of images, such as images containing people or images containing cars, and paid attention to which models performed better for those specific categories. \\ \\
    I focused on whether or not there's people, how much of the frame the train bus was, and what fraction of the train/bus was visible. \\ \\
    Model B almost always predicted ``Present''. Model A often predicted ``Absent'', even when a large vehicle was present unless it had some other vehicle to use for scale (i.e. example images 10, 13) \\ \\
    Images with buses and people, trains passing through, images without trains and buses, wide angles, close ups. \\ \\
    It was just finding correlations in features like camera angles and how much of the vehicle fits into the frame. \\
    \bottomrule
  \end{tabular}
\end{table}

\begin{table}[ht]
  \caption{Responses to ``Please describe what parts of the images/explanations contributed to your answers and how. '' for Model Selection for the GradCAM (Training only) setting}
  
  \centering
  \begin{tabular}{p{0.95\linewidth}}
    \toprule
    Response (\textit{GradCAM, Training only})\\
    \midrule
    I used the examples to try and figure out what signals each model was detecting (e.g., which areas of the image, and what those areas represented) \\ \\
    I mainly used classification results of similar images as my baseline. If the image was of a train, I took a look at which model did better with images of trains. \\ \\
    I was looking at the contrast of the color of trains relative to the background. \\ \\
    in the true label absent, even though there was no true label to reference, it generally was able to correctly classify the image as containing a vehicle when it wasn't solely the vehicle present (ie contained human/animal) where model B failed; most of the time, the images contained a vehicle and if the model A didn't classify the image as containing one, model B correctly classified it as having one but also made it so that it more mistakenly labeled the image as having a vehicle when it didn't \\ \\
    The examples that helped me the most are the ones with the ``absent'' label. I think that seeing the various examples where Model A incorrectly labeled the images ``Present'' made me decide most of the answers in the questions. \\ \\
    A simple count of correctly classified images by both models under both tabs. \\ \\
    I tried to find patterns in what was important in examples where predictions were different and by a big confidence margin, but I didn't really get much out of it. For some answers I additionally found examples with similar scenes, and directly compared the predictions. \\
    \bottomrule
  \end{tabular}
\end{table}

\begin{table}[ht]
  \caption{Responses to ``Please describe what parts of the images/explanations contributed to your answers and how. '' for Model Selection for the SmoothGrad (Training only) setting}
  
  \centering
  \begin{tabular}{p{0.95\linewidth}}
    \toprule
    Response (\textit{SmoothGrad, Training only})\\
    \midrule
    Model B looks for repeated pattern and the presence of trails whereas Model A mostly look for the main object (large vehicles). \\ \\
    The pictures where either Model A or Model B resulted in the incorrect answer and then which parts of the image each model focused on. \\ \\
    I focused on the ratio of false positives and false negatives, as well as the difference in confidence. Given a reference image, I first look at which model is more adept at handling similar scenes. Then, if they appear to be similar in performance in this regard (see: zebra photo), I look to the confidence difference in models to see which would more likely score better in one situation. \\ \\
    Model B is able to more detect vehicles correctly when they are apart of the background and generally when there are people present in the image. In contrast, Model A seems drawn to the noise created by any people in the image a well as has a stronger focus on the foreground seemingly. In examples that Model B incorrectly labels a vehicle as present, it seems that there are certain shapes that have similarities to a vehicle that may have caused this. \\ \\
    I mainly looked at the patterns between the photos that model B and model A did well in, did bad in. Using this judgment, I decided which model would do better when it came to certain images, and also considered their confidence.  \\
    \bottomrule
  \end{tabular}
\end{table}

\begin{table}[ht]
  \caption{Responses to ``Please describe what parts of the images/explanations contributed to your answers and how. '' for Model Selection for the Oracle (Training only) setting}
  
  \centering
  \begin{tabular}{p{0.95\linewidth}}
    \toprule
    Response (\textit{Oracle, Training only})\\
    \midrule
    Just looking through the sample images, I compared those with similar features like people, animals, cars, etc. \\ \\
    Whether there were humans, cars, trains/buses in pictures and where they were located. \\ \\
    I compared and contrasted the shapes of the areas being identified as important in the two models and what the models' labels were.  \\ \\
    Mostly looking at the images it was best to understand how many people were in the image and if there were tracks in the images. \\ \\
    Seeing the differences between what Model A and Model B detect as the most important part of the image was helpful to try to notice patterns for what they tend to be better at detecting. \\ \\
    There was a figure where B wasn't able to tell that the picture was a bus, even when provided with the full head. For a lot of Model A's false positives, they involved cars, which is closer to B's, which were random shades in the picture. \\
    \bottomrule
  \end{tabular}
\end{table}

\begin{table}[ht]
  \caption{Responses to ``Please describe what parts of the images/explanations contributed to your answers and how. '' for Model Selection for the GradCAM (Training and Evaluation) setting}
  
  \centering
  \begin{tabular}{p{0.95\linewidth}}
    \toprule
    Response (\textit{GradCAM, Training and Evaluation})\\
    \midrule
    I understands the model through just the right or wrong guesses, the explanations helps me see which part the model is focused more so when theres no people and no car, I can decided between A and B \\ \\
    Yellow blobs on train or blue blobs for B were in more correct positions \\ \\
    My answers were based on the examples provided (what identified what correctly) in addition to what the image contained (whether it was filled with a number of objects such as cars and people or just a single item) \\ \\
    The difference images have areas of the train/car in blue more often. \\ \\
    Mostly was looking at the difference chart and seeing which model highlighted the region where a vehicle would be present in the image. \\ \\
    Looking at the wrong predictions the models made and finding commonalities. \\ \\
    \bottomrule
  \end{tabular}
\end{table}

\begin{table}[ht]
  \caption{Responses to ``Please describe what parts of the images/explanations contributed to your answers and how. '' for Model Selection for the SmoothGrad (Training and Evaluation) setting}
  
  \centering
  \begin{tabular}{p{0.95\linewidth}}
    \toprule
    Response (\textit{SmoothGrad, Training and Evaluation})\\
    \midrule
    I looked at which parts of the images were important to each model and whether the model picked out the parts of the image with the vehicle or alternatively didn’t pick out parts of the image without the vehicle. \\ \\
    model A seems to identify as absent when its actually present \\ \\
    Since Model A is more precise in its identification/placed importance on parts of the image, it seems to be a better choice for accuracy (ex: never predicting there's a train/bus when there isn't). On the other hand, since Model B seems to notice more points (perhaps overly inclusive), it's a safer choice when you want to detect every train/bus (even when that means that it may incorrectly over-categorize objects). \\ \\
    I think my main logic was looking at important pixels, if there are a lot of important pixels then the model is so to speak confused about what's important in the image. \\ \\
    Models tended to heavily weight features of vehicles like headlights, windshields, and wheels – if these pixels were identified it was more likely to be correctly positively classified. \\ \\
    The importance map of each image is mostly similar, except that model A has a higher importance of points near people than B. The confidence map and the confidence difference were also helpful in seeing where the two models might have different behavior. \\
    \bottomrule
  \end{tabular}
\end{table}

\begin{table}[ht]
  \caption{Responses to ``Please describe what parts of the images/explanations contributed to your answers and how. '' for Model Selection for the Oracle (Training and Evaluation) setting}
  
  \centering
  \begin{tabular}{p{0.95\linewidth}}
    \toprule
    Response (\textit{Oracle, Training and Evaluation})\\
    \midrule
    The example of two people sitting at a station, and Model B assigning a Present label was one large contribution. Also, Model B assigning Absent to the image of just a truck shows it's less likely to give false positives. \\ \\
    The differences in the 4th panel and actually looking at the model present or absent. \\ \\
    The highlighted explanations in image 2 and 3 are very useful for answering how the models are different. I only used the models' labels in determining which model to use for the company. For guessing which model would have the correct prediction, I used image 2 and 3 to see if the model highlighted a train/bus to make a guess first. If model A highlights a car, I'd guess that model B is correct. Otherwise, I see if model B's explanation is rectangular on an object that is not a train/bus. If it is, then I'd guess that model A is correct. I didn't really use image 4. \\ \\
    For example in the first example photo, the section used by Model A has mainly straight lines, where the section used by B has rough edges. Same for the 8th and 11th, and 14th examples \\ \\
    In the examples Model A would always identify something, like the zebra or person, but Model B would look for anything that appears like a large vehicle. \\ \\
    Model B actually detected trains/buses, while model A consistently noticed extraneous things, like the background. \\ 
    \bottomrule
  \end{tabular}
\end{table}

\clearpage
\subsection{Counterfactual Simulation}

\begin{table}[ht]
  \caption{Responses to ``Please describe as best as possible the decision process of the model including any issues you notice. '' for Counterfactual Simulation for the no explanation setting}
  
  \centering
  \begin{tabular}{p{0.95\linewidth}}
    \toprule
    Response (\textit{No Explanation})\\
    \midrule
    The model often switched answers to absent when the front/middle of the train was covered. \\ \\
    Examining the true positive --> false negative images, it seems as though blurs that remove a significant length of a train or bus can cause false negative predictions. In cases where the blur covers a significant part of the vehicle but still leaves a ``sliver'' of the vehicle in the middle unblurred, such as the blue and yellow train or the red and beige train (which has a bit of the train visible at the front bottom), the model is still able to correctly detect the vehicle. However, in cases where an entire section from top to bottom of a vehicle is obscured, such as in the green and yellow train, the model may fail to detect the vehicle. \\ \\
    It doesn't seem to introduce a present judgement when the original was absent \\ \\
    It likely looks for a vehicle-like shape in the image (only looking at the outside of the vehicle) and makes its prediction based on how much of the screen the vehicle takes up. It has trouble detecting certain vehicles at all (i.e. when the color is similar to the background) and fails when the blur covers up most of the vehicle/makes the vehicle look smaller (i.e. by covering up the middle section). \\ \\
    It is more likely that the model makes the classification of ``absent'', if the model's prediction on the unblurred image is labeled ``absent'' too and vice versa.  \\ \\
    Using those two details, I noticed, I avoided marking an image ``Positive'' when the original had an ``Absent'' label and otherwise made judgement calls about how much of the vehicle was being obscured by blurring. \\
    \bottomrule
  \end{tabular}
\end{table}

\begin{table}[ht]
  \caption{Responses to ``Please describe as best as possible the decision process of the model including any issues you notice. '' for Counterfactual Simulation for the GradCAM (Training only) setting}
  
  \centering
  \begin{tabular}{p{0.95\linewidth}}
    \toprule
    Response (\textit{GradCAM, Training only})\\
    \midrule
    It seems like the model uses key features of the trains (e.g., the windows) to detect a train, and when those are missing, it fails to detect it. It also appears to detect only one bus/car/etc. in a photo with many, so if one is hidden, it may miss the other. \\ \\
    It looks for streaks of similar colour that form a pattern, like the side of a bus or train. If it finds them, it is probably present. However, large trucks start looking like present when they should be absent. \\ \\
    I am not very knowledgable of this but I think the model looks for specific pixels associated with the main subject of the photo then detects if that is or is not what they are looking for. In this case, if the box highlights a real object then I guessed present. \\ \\
    sometimes the lack of or existence of features were used (aka, which part of the image was used to make the classification was not consistent) \\ \\
    I think that the model looks for certain shapes of objects that are vehicular. The main problem is that if a vehicle is at an angle that's outside of a 45 degree view of it or if the blur covers most of the vehicle, then the model can't see that the vehicle IS a big vehicle based on something ``smaller'' like a license plate or a ``big blog'' that we can see contains wheels. \\ \\
    For most images I picked absent, except for when I believed the blur was not covering all the most informative areas. \\ \\
    The model seems to look for some signs of a large vehicle (boxy, long lines with parallax indicating depth, maybe?). This seems to cause trains that are in the distance, or with funny shapes present that break up this silhouette, to be missed. It also misses interiors, and views of large vehicles close up. \\
    \bottomrule
  \end{tabular}
\end{table}

\begin{table}[ht]
  \caption{Responses to ``Please describe as best as possible the decision process of the model including any issues you notice. '' for Counterfactual Simulation for the SmoothGrad (Training only) setting}
  
  \centering
  \begin{tabular}{p{0.95\linewidth}}
    \toprule
    Response (\textit{SmoothGrad, Training only})\\
    \midrule
    The model only relies on part of the object \\ \\
    If most of the train is blurred, the model will incorrectly determine that there is no train. It also does not seem to be able to take parts of the train/bus that is far compared to the point of view of the camera into account. \\ \\
    I noticed that the model focuses a lot on the edges/contours of the trains and buses, including headlights and roof shapes. This is a bit of an issue as all vehicles have headlights, and many vehicles share the rounded roof style of many buses and trains. Additionally, as the creators want the model to predict correctly with smudged lenses, contours may be altered, blurred out, and otherwise hard or impossible to see. \\ \\
    The model seems unable to identify things in the blurred region and relies on the remaining picture outside of it. \\ \\
    It seems like the model struggles if there is only the front of the train, but does fine if there's a tail. If the blur image has a blur with a tail, it'll state it is present. \\
    \bottomrule
  \end{tabular}
\end{table}

\begin{table}[ht]
  \caption{Responses to ``Please describe as best as possible the decision process of the model including any issues you notice. '' for Counterfactual Simulation for the Oracle (Training only) setting}
  
  \centering
  \begin{tabular}{p{0.95\linewidth}}
    \toprule
    Response (\textit{Oracle, Training only})\\
    \midrule
    To be honest, I don't really know what the decision process is. I think if it clearly sees a train, it will identify the image as present, but sometimes the background is confusing. \\ \\
    If front of bus/train is recognized, follow to form entire object. Doesn't work for inside of train/bus. \\ \\
    It took me quite a while to realize what the exercise was asking, then to understand if the model grouped pixels and assigned importance to where the pixels were located on the image, or if it's just if the blur covered the distinguishing characteristics of a large vehicle.  \\ \\
    The models. attempt to look for large rectangles that denote vehicles. \\ \\
    The model seems to be better at detecting the fronts of vehicles, especially from the outside and when most/all of the vehicle is present. It's also better when the background isn't busy. It seems to be worse at detecting the insides of vehicles and the back cars of trains. \\ \\
    Similar to how I learn based off of the examples, the model is figuring out patterns between the present vs. absent images. When a new image comes up that has similar features, comparing it to previous examples will influence the final decision of the model. \\
    \bottomrule
  \end{tabular}
\end{table}

\begin{table}[ht]
  \caption{Responses to ``Please describe as best as possible the decision process of the model including any issues you notice. '' for Counterfactual Simulation for the GradCAM (Training and Evaluation) setting}
  
  \centering
  \begin{tabular}{p{0.95\linewidth}}
    \toprule
    Response (\textit{GradCAM, Training and Evaluation})\\
    \midrule
    Basically if the important part is blurred it predicts absent. \\ \\
    the model pays more attention to front and neglects sides or shrunk back parts due to perspective \\ \\
    The main issue I noticed was that the results might become incorrectly skewed based upon what the original model identifies. It is my understanding that the model identifies key distinct featuers in the image, if the identified features align with what has been defined as a train or truck, then it makes its decision accordingly.  \\ \\
    If the model focuses on certain areas that aren't visible because of blurring, it has difficulty adjusting. \\ \\
    The model seems to focus on contextual details such as road signs and train tracks to determine whether or not a vehicle is present. \\ \\
    Smaller vehicles are easily mistaken to be buses/trains. The model also fails to recognize trains when the image lacks large, discernable windows. \\
    \bottomrule
  \end{tabular}
\end{table}

\begin{table}[ht]
  \caption{Responses to ``Please describe as best as possible the decision process of the model including any issues you notice. '' for Counterfactual Simulation for the SmoothGrad (Training and Evaluation) setting}
  
  \centering
  \begin{tabular}{p{0.95\linewidth}}
    \toprule
    Response (\textit{SmoothGrad, Training and Evaluation})\\
    \midrule
    Much of the image was blurred so it was a hard task. \\ \\
    the model decides based on the bright parts of the image  \\ \\
    Looking at the examples, it seems like the model attempts to look at outlines or high concentrations of 'activity' in the image, and categorizes from there. Specifically looking at examples where the original and blurred resulted in different conclusions though, it seems like the model has some issues when the blur causes these distinct outlines/edges, or activity in the image, to become unclear, leading to inaccurate results. \\ \\
    Does it always detect absent when it detected absent in original as well? Sometimes this can be bad because what if it paid attention to wrong features and blurring out could potentially help? \\ \\
    Same logic as previous – it is surprisingly resilient to blurring though. \\ \\
    The model is looking for rectangular/defined features that are associated with these vehicles (corners of buses, well defined windows, etc) so in situation where key features are blurred the model switches a present prediction to absent. This blurring doesn't affect the model the other way (changing a prediction from absent to present), which contributes to why I think the model uses sharp features to predict present. \\
    \bottomrule
  \end{tabular}
\end{table}

\begin{table}[ht]
  \caption{Responses to ``Please describe as best as possible the decision process of the model including any issues you notice. '' for Counterfactual Simulation for the Oracle (Training and Evaluation) setting}
  
  \centering
  \begin{tabular}{p{0.95\linewidth}}
    \toprule
    Response (\textit{Oracle, Training and Evaluation})\\
    \midrule
    The model typically looks for the train or bus in images, and when it sees that, it assigned a Present label. When the large vehicle is blurred, then the model possibly returns an Absent label instead, even if there are parts that can see be seen. \\ \\
    Windows, train wheels, rectangular parts and backgrounds. \\ \\
    The model seems to detect a horizontal long-ish rectangles as ``Present''. As long as the blur box preserves this shape more or less, the model seems to still correctly predict ``Present''. It also has a tendency to predict cars as ``Present''. If the blur box is able to cover most of the cars such that the cars don't look like cars, then it won't falsely predict ``Present''. (Also not sure if this is intended, but there is an image with true label ``Present'' in the Absent tab. It's the same image as the example in the instructions.)  \\ \\
    It seems to look for roughly train shaped items, and it if detects them it says Present. However it seems to often miss trains, especially when the colors are strange or similar to the background color \\ \\
    The model tries to identify longer horizontal colors that stick out from the background and if they are different enough it classifies it as a large vehicle. I noticed the model also seems to look at the tracks in front of the train in some cases, and this could maybe be an issue if it was given a model of railroad tracks with no train on them. \\ \\
    if it detects the train/bus, it says it saw it. \\
    \bottomrule
  \end{tabular}
\end{table}

\begin{table}[ht]
  \caption{Responses to ``Please describe what parts of the images/explanations contributed to your answers and how.'' for Counterfactual Simulation for the no explanation setting}
  
  \centering
  \begin{tabular}{p{0.95\linewidth}}
    \toprule
    Response (\textit{No Explanation})\\
    \midrule
    whether the front/middle of a train/bus was covered, what exactly was covered, etc \\ \\
    When a positive was predicted originally, I focused on what parts of the image were blurred, depending on the angle and size of the object. Since there were no instances of an original absence prediction becoming a modified presence prediction, if the model originally predicted there to be no train or bus, I believe the model would still predict there to be no train or bus after the modification. If the majority of the vehicle was obscured, I predicted that the model would believe the modified image to not have a vehicle in it. \\ \\
    The bluring seems to confuse the model in particular when it erases specific features like wheels/train wheels/train tracks. \\ \\
    In general, if most of the large vehicle was obscured, I tended to answer that the model would switch from Present to Absent. Furthermore, if the model's answer was already Absent, I answered that it would stay predicting Absent as none of the example images showed a change from Present to Absent. \\ \\
    Different close ups, wide shots. Where is the image blurred, e.g. more in the middle or in the corner. \\ \\
    - The model never goes from labeling an image from ``Absent'' to ``Present'' after blur is applied. - If most or all of the vehicle is obscured, it downgrades from ``Present'' to ``Absent.'' \\
    \bottomrule
  \end{tabular}
\end{table}

\begin{table}[ht]
  \caption{Responses to ``Please describe what parts of the images/explanations contributed to your answers and how.'' for Counterfactual Simulation for the GradCAM (Training only) setting}
  
  \centering
  \begin{tabular}{p{0.95\linewidth}}
    \toprule
    Response (\textit{GradCAM, Training only})\\
    \midrule
    I tried to find similar images in the examples to see when blurring changed the model's results. \\ \\
    The model appeared to be mainly detecting sides of vehicles, so if those were blurred, it would become absent. There were no examples of absent turning to present when blurring, so I assumed that would remain the case. \\ \\
    I was looking at what the box covered and if it is one object. \\ \\
    looking at similar images in the examples and seeing which parts of the images were deemed more important  \\ \\
    When deciding on an answer, I was mainly seeing if there were any vehicle-shaped rectangles in the image. If there was still such a shape, I'd mark ``Present''. \\ \\
    When the image is blurred in an area with high importance, the model seems to be more likely to classify it as absent. \\ \\
    I looked at what parts of images being blurred out contributed to changes in decisions. \\
    \bottomrule
  \end{tabular}
\end{table}

\begin{table}[ht]
  \caption{Responses to ``Please describe what parts of the images/explanations contributed to your answers and how.'' for Counterfactual Simulation for the SmoothGrad (Training only) setting}
  
  \centering
  \begin{tabular}{p{0.95\linewidth}}
    \toprule
    Response (\textit{SmoothGrad, Training only})\\
    \midrule
    which portion is covered and the model's decision in similar examples \\ \\
    How much of the object of interest is obstructed from the blur filter.  \\ \\
    To determine what the model was focusing on, I used the pixel heatmap with the purple to orange gradient that shows which pixels are most important to the model, also giving a hint as to what components of the objects the model finds most interesting. \\ \\
    Partial obfuscation of the vehicle seems to be okay as there is still identifiable components of it still visible. \\ \\
    Like before, I simply tried to match patterns, as I tried to find similar aspects and see how the model treated those aspects.  \\
    \bottomrule
  \end{tabular}
\end{table}

\begin{table}[ht]
  \caption{Responses to ``Please describe what parts of the images/explanations contributed to your answers and how.'' for Counterfactual Simulation for the Oracle (Training only) setting}
  
  \centering
  \begin{tabular}{p{0.95\linewidth}}
    \toprule
    Response (\textit{Oracle, Training only})\\
    \midrule
    I looked a lot at the background images. \\ \\
    Whether front of train is blurred and how big the train. Whether inside or outside of the train. \\ \\
    I looked for where the blurred parts were within the photo and compared those with the example photos in the true labels  to see what the model would predict if that region is blurred. \\ \\
    To me it was most important what part of the image was covered up. If the transit part of the image was covered up then it was most likely to not be identified by the models. \\ \\
    I noticed that if the blurred-out part of the modified image completely covered the detected part of the original image, then the modified image was marked ``absent'' even though the original image was marked ``present''. Additionally, there were no instances of true label absent, original absent, and modified present, so my answers reflect this as well. \\ \\
    The model is good at identifying autos, which unfortunately caused a lot of false positives. It's also not good at identifying trains when at least a third of the picture is blurred. \\
    \bottomrule
  \end{tabular}
\end{table}

\begin{table}[ht]
  \caption{Responses to ``Please describe what parts of the images/explanations contributed to your answers and how.'' for Counterfactual Simulation for the GradCAM (Training and Evaluation) setting}
  
  \centering
  \begin{tabular}{p{0.95\linewidth}}
    \toprule
    Response (\textit{GradCAM, Training and Evaluation})\\
    \midrule
    See if the important part is blurred \\ \\
    look at is front or large section blurred \\ \\
    The main parts of the image that contributed to my answer was based upon what the original model identified as important. If that contained the vehicle or key identifying features, then I decided on the modified model accordingly (if the orignal model identified only the blurred section as important then I made my decision as absent). \\ \\
    If blurred areas were more yellow-y, it seems like it would change the answer if that region were removed. \\ \\
    If the image was blurred where the model marked high importance, I think it might mess with the ability to detect anything present. \\ \\
    When only parts of the vehicle are blurred, and the original prediction is present, the modified prediction is usually also present. But when the entire vehicle is blurred, the modified prediction could change.  \\
    \bottomrule
  \end{tabular}
\end{table}

\begin{table}[ht]
  \caption{Responses to ``Please describe what parts of the images/explanations contributed to your answers and how.'' for Counterfactual Simulation for the SmoothGrad (Training and Evaluation) setting}
  
  \centering
  \begin{tabular}{p{0.95\linewidth}}
    \toprule
    Response (\textit{SmoothGrad, Training and Evaluation})\\
    \midrule
    I looked at which parts were blurry and compared it to which parts the model deemed important to see how it would predict.  \\ \\
    if the blur cover the bright parts of the image then it can change the label \\ \\
    I think that it was really looking at what points the model placed importance on in the original image, and seeing how those points were altered in the appearance of the blurred image, then deciding whether or not those originally important characteristics were still clear/distinct in the blurred image (corresponding to whether or not the blurred image would or wouldn't have the same conclusion as the original). \\ \\
    If important pixels were not that yellow and they were blurred, the model now cannot detect anymore. If the yellow pixels were very bright and covered it still can tell. \\ \\
    If important pixels for classification were blurred / obscured, I expected that the model would have a much more difficult time correctly classifying those images. \\ \\
    The importance map helped me understand that the model looks to points of definition or sharp change (corners, borders, etc). This also made it easier to understand if the blurred region overlapped where the importance points were. \\
    \bottomrule
  \end{tabular}
\end{table}

\begin{table}[ht]
  \caption{Responses to ``Please describe what parts of the images/explanations contributed to your answers and how.'' for Counterfactual Simulation for the Oracle (Training and Evaluation) setting}
  
  \centering
  \begin{tabular}{p{0.95\linewidth}}
    \toprule
    Response (\textit{Oracle, Training and Evaluation})\\
    \midrule
    I looked at the example images and noticed that when the modified image blurred the train/bus, the model was going to assign an Absent label after. If the original was Absent, the modified was Absent as well. \\ \\
    Seeing which parts were not blurred out such as railroad tracks or part of the train stations. \\ \\
    If the original label is ``Absent'', I just guess the model will say ``Absent'' regardless of the blur. Otherwise, I used the heatmap to see what the model is focusing on and the modified image to see where the blur box is. If the blur box is put on about ~90\% of the yellow highlight, then I'd guess that the model predicts ``Absent'' instead of ``Present''. 
     \\ \\
     When there is a clear shape used by the model it says Present, and when the heatmap shows no clear shapes it is Absent \\ \\
     I looked at what sort of patterns the model was picking out, and how much the blur impacted those patterns. If the blur made it hard to distinguish between the border of the train or not I would assume the model would fail to identify the train. \\ \\
     obscuring the majority of the train will make you unlikely to be able to identify. this is the same for the models. \\
    \bottomrule
  \end{tabular}
\end{table}


\end{document}